\documentclass[11pt,letterpaper]{article}

\usepackage[margin=1in]{geometry}
\usepackage[T1]{fontenc}
\usepackage{lmodern}
\usepackage{amsmath,amssymb}
\usepackage{graphicx}
\usepackage{booktabs}
\usepackage{array}
\usepackage{tabularx}
\usepackage{multirow}
\usepackage{enumitem}
\usepackage{microtype}
\usepackage{caption}
\usepackage{placeins}
\usepackage{natbib}
\setcitestyle{authoryear,round}

\usepackage{xspace}
\usepackage[table]{xcolor}
\usepackage{tikz}
\usetikzlibrary{positioning,shapes.geometric,arrows.meta,fit,backgrounds,calc}
\usepackage[hidelinks]{hyperref}


\definecolor{tlpurple}{HTML}{6D28D9}
\definecolor{tlpurpledeep}{HTML}{211832}
\definecolor{tlteal}{HTML}{0F766E}
\definecolor{tlteallight}{HTML}{ECFDF5}
\definecolor{tlpurplelight}{HTML}{F5EDFF}
\definecolor{tlslate}{HTML}{0F172A}
\definecolor{tlmuted}{HTML}{4B5563}

\hypersetup{
  pdftitle={Posterior Twins: Distributional Behavioral Simulation for Enterprise Decisions},
  pdfauthor={Ankit Das},
  pdfsubject={Distributional behavioral simulation and digital twins},
  pdfkeywords={behavioral simulation, digital twins, distributional fidelity, Wasserstein distance, enterprise AI}
}

\newcommand{\Wone}{W_{1}}
\newcommand{\acct}{\mathrm{Acc}_{\mathrm{theoretical}}}
\newcommand{\keywords}[1]{\par\noindent\textbf{Keywords:} #1}
\newcommand{\tlmetric}[3]{%
  \begin{center}
    {\footnotesize\bfseries #1}\par\vspace{2pt}
    {\Large\bfseries #2}\par\vspace{2pt}
    {\footnotesize #3}
  \end{center}
}
\newenvironment{tlhighlight}
  {\begin{quote}\begin{minipage}{\linewidth}\small\noindent\rule{\linewidth}{0.4pt}\par\medskip}
  {\par\medskip\noindent\rule{\linewidth}{0.4pt}\end{minipage}\end{quote}}
\newenvironment{tlcallout}[1]
  {\begin{quote}\small\noindent\textbf{#1.}\ }
  {\end{quote}}
\newenvironment{tlevidence}[1]
  {\begin{quote}\small\noindent\textbf{#1.}\ }
  {\end{quote}}
\newenvironment{tlscope}[1]
  {\begin{quote}\small\noindent\textbf{#1.}\ }
  {\end{quote}}
\makeatletter
\renewcommand\paragraph{\@startsection{paragraph}{4}{\z@}%
  {1.25ex plus 0.2ex minus 0.2ex}%
  {0.7ex}%
  {\normalfont\normalsize\bfseries}}
\makeatother

\title{Posterior Twins: Distributional Behavioral Simulation for Enterprise Decisions}
\author{Ankit Das\\Twinning Labs, Inc.\\\texttt{ankit@twinninglabs.com}}
\date{May 2026}

\begin{document}
\maketitle

\begin{abstract}
Enterprise behavioral simulation requires more than producing a plausible response. Many decisions depend on the shape of a population under a proposed action: which segments accept, defect, hesitate, or move into risk-sensitive states. This paper introduces Posterior Twins, a memory-grounded digital-twin approach that represents likely behavior as an updated distribution under a specific decision context. We evaluate a family of Twinning Labs behavioral-model operating points on a 226-example held-out behavioral-response benchmark and report both modal accuracy and Wasserstein-1 distance. The results show that modal accuracy and distributional fidelity identify different operating regimes. TL-Twin Alpha achieves the lowest observed Wasserstein-1 distance in the reported result set ($\Wone=1.16$), while TL-Twin Delta and TL-Twin Gamma provide balanced operating points near the modal-accuracy frontier. The paper frames these results as a systems result: governed memory, behavioral model routing, scenario orchestration, distributional aggregation, and auditability are necessary for turning simulated behavior into reusable enterprise decision evidence.
\end{abstract}


%====================================================================
\section{Introduction}
\label{sec:exec-summary}

Enterprise behavioral simulation goes beyond response generation. A basic generator produces a plausible output. An enterprise simulator reproduces the shape of a population under a decision: who accepts, who defects, who hesitates, which segments move, and how much uncertainty remains. The central claim of this paper is simple: \textbf{modal accuracy and distributional fidelity measure different capabilities}. Systems that look similar on point accuracy can behave very differently when judged by the population distribution they produce.

A \textbf{Posterior Twin} is a memory-grounded digital twin represented as an updated distribution over likely behavior under a specific decision context. In enterprise terms, it uses governed customer evidence to estimate what a customer, account, or segment is likely to choose, say, or do, and how uncertain that estimate is. Because the output is a distribution, it can be measured, compared, rerun, and audited.

\begin{tlhighlight}
\textbf{What this paper establishes.}
\begin{itemize}[leftmargin=1.2em,topsep=2pt,itemsep=2pt]
  \item \textbf{Twinning Labs' TL-Twin Alpha sets the distributional-fidelity point.} TL-Twin Alpha achieves the lowest observed Wasserstein-1 distance in the reported result set ($\Wone=1.16$), with modal accuracy of $0.705$.
  \item \textbf{The system also provides balanced TL operating points near the frontier.} TL-Twin Delta reaches modal accuracy of $0.750$ with $\Wone=2.36$, while TL-Twin Gamma reaches modal accuracy of $0.749$ with $\Wone=2.28$.
  \item \textbf{The result is a routable operating frontier.} The Posterior Twin system routes each scenario to the operating point that fits its dynamics: population movement, calibrated behavior, scenario stability, or direct execution.
\end{itemize}
\end{tlhighlight}

\begin{center}
\begin{minipage}[t]{0.31\linewidth}
\vspace{0pt}%
\tlmetric{SHAPE}{1.16}{Lowest observed $\Wone$ in the reported result set.}
\end{minipage}\hfill
\begin{minipage}[t]{0.31\linewidth}
\vspace{0pt}%
\tlmetric{BALANCE}{0.750}{TL-Twin Delta reaches $0.750$; TL-Twin Gamma reaches $\Wone=2.28$ at $0.749$.}
\end{minipage}\hfill
\begin{minipage}[t]{0.31\linewidth}
\vspace{0pt}%
\tlmetric{SYSTEM}{Digital Twins}{Memory-grounded twins acting inside scenario environments.}
\end{minipage}
\end{center}

\begin{tlevidence}{Evaluation set}
The reported result set uses the comparable 226-example held-out evaluation. It includes the TL operating points shown in the system architecture and every frontier-model row evaluated on the same benchmark. The table reports point estimates for every row under the same convention.
\end{tlevidence}

The enterprise consequence is straightforward: a usable simulation system needs maintained memory, memory-grounded twins, scenario orchestration, structured decision artifacts, distributional aggregation, and a governance boundary around customer evidence. Posterior Twins close that loop. The TL model frontier supplies behavior, the Memory Layer supplies persistent evidence, and the Simulation Engine turns both into repeatable, auditable decision runs.

Different decisions create different operating requirements. A campaign simulation may depend on segment movement. A pricing simulation may depend on population-shape fidelity because willingness-to-pay, downgrade risk, and rejection tails drive the economics. A product simulation may need both. The benchmark is therefore a system-design result: Twinning Labs exposes operating points that the Simulation Engine can route, rerun, compare, and audit as decision evidence.

%====================================================================
\section{The Category Problem}
\label{sec:category}

This is not one product category with different wrappers. It is a cluster of systems approaching the same enterprise problem from different angles: synthetic audiences, AI-moderated research, digital-twin systems, multi-agent population simulators, and customer-data decisioning products.

These systems differ in product surface. Some expose conversational personas. Some generate respondent-level synthetic survey data. Some model interacting agents. Some optimize live customer actions from observed outcomes. The technical boundary that matters is the quality of the decision evidence: whether the system gives the decision-maker a defensible response distribution under a specified decision context.

The distinction matters because enterprise decisions are allocation decisions. A marketing team allocates budget across segments. A pricing team allocates margin across willingness-to-pay curves. A product team allocates roadmap weight across user groups that may react differently to the same launch. A risk team allocates reserve and review capacity across tail exposure as well as median expectation.

\begin{tlcallout}{The category error}
A polished persona, modal respondent, or next-best-action can be useful. Enterprise decisions that depend on heterogeneity, calibration, uncertainty, and tail behavior need an evaluable behavioral output: sometimes a response distribution that can be compared to empirical distributions, sometimes a generated output, action, or behavioral trace that can be reviewed against the twin's memory and scenario state.
\end{tlcallout}

A useful enterprise simulator therefore has to preserve four properties at once. First, for choice-structured tasks, it has to preserve \textbf{decision direction}: the simulated movement typically tracks what real people are most likely to do under the same task structure. Second, for open-ended tasks, it has to preserve \textbf{behavioral generation}: the generated output, action, or behavioral trace follows from the twin's evidence state rather than from persona-only role-play. Third, across both task types, it has to preserve \textbf{population shape}: the simulator keeps probability mass in the right behavioral states when the output is scored as a distribution. Fourth, it has to preserve \textbf{decision trace}: the team can rerun a scenario, compare it to another scenario, and understand which evidence state produced the output.

These requirements are why fluency, persona richness, and single-score benchmarks are incomplete measures of enterprise simulation. Aggregate distribution, tail behavior, and segment movement determine whether simulated behavior becomes useful decision evidence. A distribution-first operating point becomes the right tool when the decision depends on risk, sensitivity, or segment movement rather than a single winning outcome.

The literature points in this direction. Silicon-sampling work showed that language models encode useful behavioral priors~\citep{argyle2023,horton2023}. Prompt-conditioning studies exposed how sensitive those priors can be under demographic and political framing~\citep{santurkar2023whose}. Generative-agent work made simulated social dynamics visible but also sharpened the need for measurement~\citep{park2023generative}. Cognitive foundation model and behavioral-benchmark work has moved the field toward distributional comparison, repeated measurement, and explicit simulation protocols~\citep{binz2024centaur,huang2025socrates,simbench2025,twin2k2025}. The economic and behavioral traditions have made the same point for decades: decisions live in distributions rather than stories~\citep{samuelson1938,mcfadden1974,benakiva1985,train2009,fishbein1975,ajzen1991}.

The practical conclusion is that enterprise behavioral simulation should be evaluated as a decision system: distributional fidelity, modal accuracy, data boundary, cost envelope, repeatability, and auditability.

%====================================================================
\section{The Twinning Labs Approach}
\label{sec:approach}

Twinning Labs builds Posterior Twins as a complete behavioral simulation system. The TL model family supplies the operating frontier measured in this paper. The Memory Layer connects governed enterprise evidence across customer, product, research, commercial, and outcome systems, then preserves that evidence as stable twin context. Together, the TL operating point and the Memory Layer instantiate the digital twin: a memory-grounded behavioral actor that can choose, respond, reason, or act under a decision context. The Simulation Engine creates scenario environments around those twins, maps outcomes to the scenario contract, aggregates distributions or structured generative artifacts, and exposes the result as an auditable decision object. The word ``posterior'' is deliberate: a twin is an updated behavioral state under a given decision context.

\begin{center}
\centering
\begin{tikzpicture}[
  font=\sffamily\small,
  layer/.style={
    rectangle,
    rounded corners=3pt,
    minimum width=15.0cm,
    minimum height=1.32cm,
    align=center,
    line width=0.55pt
  },
  memory/.style={layer, fill=tlteallight, draw=tlteal!45},
  simulation/.style={layer, fill=tlpurple!8, draw=tlpurple!45},
  twin/.style={layer, fill=tlteal!12, draw=tlteal!55},
  model/.style={layer, fill=tlpurple!16, draw=tlpurple!60},
  trust/.style={layer, fill=tlpurpledeep!8, draw=tlpurpledeep!65},
  boundary/.style={
    rectangle,
    rounded corners=8pt,
    inner sep=10pt,
    draw=tlpurpledeep!35,
    line width=0.85pt,
    dashed,
  }
]
\node[memory] (memory) at (0,0) {%
  \textbf{Memory Layer}\\[2pt]
  {\footnotesize Connected first-party evidence state $\cdot$ governed context $\cdot$ historical observations}};
\node[model, above=0.28cm of memory] (model) {%
  \textbf{TL model operating frontier}\\[2pt]
  {\footnotesize Population movement $\cdot$ calibrated behavior $\cdot$ scenario simulation $\cdot$ direct execution}};
\node[twin, above=0.28cm of model] (twin) {%
  \textbf{Digital Twin Layer}\\[2pt]
  {\footnotesize Memory-conditioned behavioral actors $\cdot$ choices $\cdot$ generated actions $\cdot$ traces}};
\node[simulation, above=0.28cm of twin] (simulation) {%
  \textbf{Simulation Engine}\\[2pt]
  {\footnotesize Scenario environments $\cdot$ population runs $\cdot$ outcome mapping $\cdot$ distributional aggregation}};
\node[trust, above=0.28cm of simulation] (trust) {%
  \textbf{Trust, integration, and governance boundary}\\[2pt]
  {\footnotesize Customer-cloud option $\cdot$ auditability $\cdot$ governed access $\cdot$ workflow interfaces}};

\begin{scope}[on background layer]
\node[boundary, fit=(memory) (model) (twin) (simulation) (trust), label={[font=\sffamily\footnotesize\itshape, text=tlpurpledeep!90]above:Complete Posterior Twin system}] {};
\end{scope}
\end{tikzpicture}

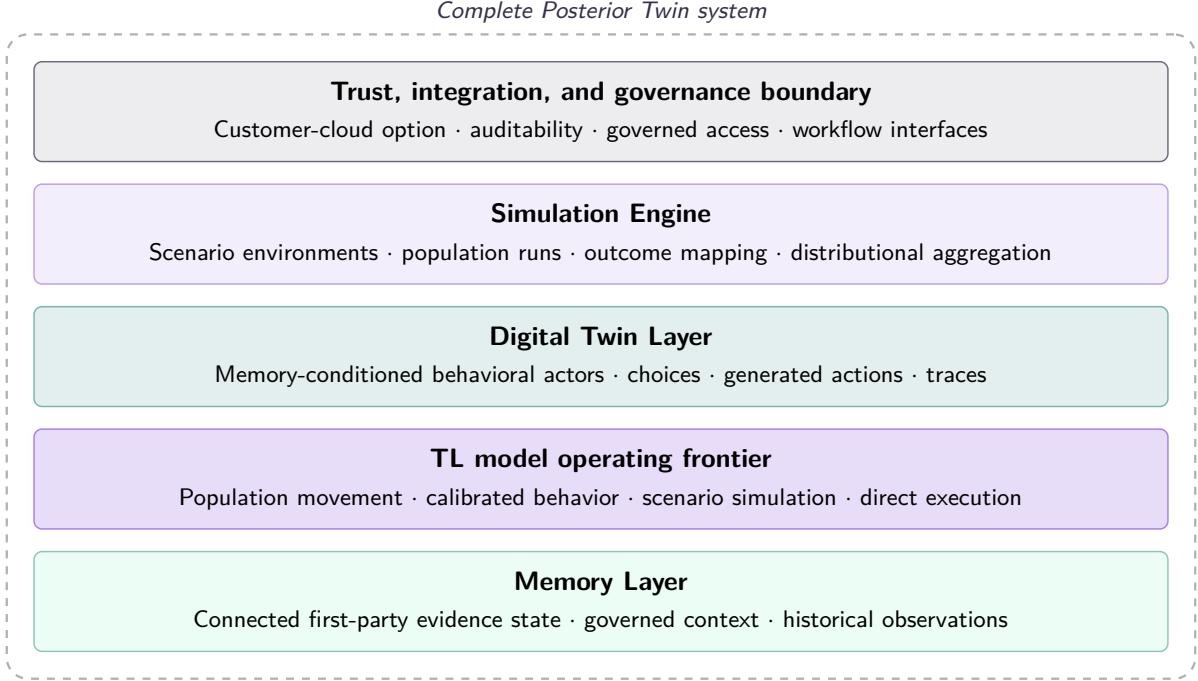
\captionof{figure}{Posterior Twins as a memory-grounded behavioral simulation system. The benchmark in this paper measures the TL model operating frontier inside the system; the Memory Layer supplies governed enterprise context from connected evidence systems, the Digital Twin Layer turns model behavior and memory into actors, and the Simulation Engine makes those actors operational through scenario orchestration and measurement inside the governed system boundary.}
\label{fig:architecture}
\end{center}

The approach has four design commitments.

\paragraph{1. Ground digital twins in persistent memory} The Memory Layer keeps a continuously maintained first-party evidence state connected to the systems where enterprise context is created: customer and account records, product usage, research, campaign and commercial history, support and customer-success context, transaction activity, and observed outcomes. It normalizes those signals into governed behavioral context over time, giving each twin durable memory rather than a disposable persona prompt.

\paragraph{2. Make TL models the behavioral core} The TL model family exposes multiple operating points. When those operating points are conditioned on the Memory Layer, they become digital twins that can make choices, generate responses, produce actions, or return behavioral traces under a decision context. A pricing simulation and a campaign-routing simulation may value different parts of the accuracy/fidelity frontier. Posterior Twins treat those operating points as system resources that the Simulation Engine can route automatically based on scenario dynamics, evidence state, and decision objective.

\paragraph{3. Let the Simulation Engine create the decision environment} The Simulation Engine defines the scenario environment, executes population-scale runs, maps outcomes onto the decision contract, aggregates distributions, compares outcomes, and records the run artifact. Evaluation is built into the engine that turns twin behavior into decision evidence.

\paragraph{4. Govern the system where enterprise evidence lives} Customer-cloud deployment is the trust boundary around the system. It keeps governed evidence, simulation records, and operating workflows in the environment where they already live.

\paragraph{The TL-Twin model family} The TL-Twin rows are Twinning Labs' in-house behavioral-model operating points: different calibrated configurations exposed through the same Posterior Twin system. \textbf{TL-Twin Alpha} is the population-movement model, optimized for distributional fidelity across customer states. \textbf{TL-Twin Beta} is the calibrated behavior model. \textbf{TL-Twin Gamma} combines multiple TL behavioral runs to stabilize scenario forecasts across decision alternatives. \textbf{TL-Twin Delta} is the direct scenario model for lower-latency decision runs. The names are shorthand for distinct behavioral roles inside one system.

\paragraph{Memory as the compounding asset} A one-off synthetic response discards context after the output is generated. A memory-grounded simulator keeps the evidence state available for future runs across customer, product, commercial, research, and outcome systems. That persistence is what lets a company compare today's pricing scenario to last quarter's research, or rerun a campaign concept after new customer evidence arrives.

\paragraph{Digital twin grounding as deployment lift} The benchmark evaluation scores every row against the same benchmark context. In deployment, the Memory Layer gives each twin a continuously maintained first-party evidence state assembled from connected enterprise systems and governed customer evidence. The Simulation Engine then places those memory-grounded twins inside scenario environments that mirror the real-world decision context. This enrichment is designed to strengthen grounding, modal selection, segment movement, and distributional fidelity because the twin is conditioned on the customer's actual behavioral evidence.

\paragraph{Simulation as measurement infrastructure} The Simulation Engine is the layer that makes the digital twin population usable for decisions. It turns a decision into a scenario portfolio, executes population-scale runs, maps choice-structured outcomes onto the task scale, aggregates distributions where the task has a response scale, and returns structured behavioral artifacts where the task is open-ended. This is the difference between generated output and decision infrastructure.

\begin{tlevidence}{Why the benchmark matters}
The $1.16$ $\Wone$ result demonstrates that a TL operating point can preserve population shape on the comparable held-out evaluation better than the compared rows. In the complete Posterior Twin system, that capability becomes an automatically routable behavior inside memory-grounded simulation workflows.
\end{tlevidence}

%====================================================================
\section{Evaluation Method}
\label{sec:method}

We evaluate on a held-out split from a behavioral-response benchmark derived from approximately $210$ real-world decision studies, approximately $2.9$M responses, and approximately $400$K participants~\citep{huang2025socrates}. The result set uses $226$ held-out examples.

The paper uses two headline metrics because enterprise simulation has to support two different decision questions: whether the system identified the most likely human behavior on a choice-structured task, and whether it reproduced the full population shape.

Each evaluation item has an empirical human response distribution over an ordered theoretical scale. The simulator produces a response that can be mapped onto that same scale. This choice-structured evaluation is why modal accuracy and $\Wone$ are appropriate headline metrics here. For open-ended simulations, a memory-grounded digital twin can generate a behavioral trace, response, or action from its evidence state, and the Simulation Engine maps that result into the artifact the enterprise workflow needs.

We compare the benchmark outputs to two human targets: the empirical modal response and the full empirical response distribution. This pairing is important because the two targets correspond to different enterprise decisions. Winner-selection decisions emphasize the modal response. Forecasting, allocation, pricing, and risk decisions require the distribution.

\paragraph{Benchmark evaluation versus memory-enriched deployment} The reported numbers measure the TL operating frontier under a comparable held-out evaluation where every row is scored against the same evidence context. Production deployments enrich that frontier with first-party memory from connected customer, product, commercial, research, and outcome systems. That richer evidence state is designed to strengthen grounding, modal selection, segment movement, and distributional fidelity in production workflows, with partner backtests quantifying deployment lift in the customer's operating context.

\subsection{Evaluation Scope and Availability}
\label{sec:scope-availability}

The reported results use a comparable $226$-example held-out evaluation derived from a behavioral-response benchmark. Every row in the result table is scored under the same benchmark scale, target convention, and aggregation procedure. We report point estimates for modal accuracy and Wasserstein-1 distance because they measure complementary decision properties: agreement with the empirical modal response and fidelity to the full empirical response distribution.

The TL-Twin rows are Twinning Labs behavioral-model operating points exposed through the Posterior Twin system. The benchmark evaluation uses the benchmark item context and empirical response distributions; it is separate from customer-cloud deployments that enrich scenario runs with governed first-party memory. This separation keeps the comparison focused on the operating frontier while preserving the deployment architecture that makes Posterior Twins useful in enterprise settings.

The paper provides the evaluation definitions, metric formulas, table-level point estimates, and rendered figures needed to interpret the reported operating frontier. Twinning Labs retains the implementation details of the TL-Twin operating points, routing logic, Memory Layer integrations, and production simulation workflows as proprietary system assets. This boundary is intentional: the paper contributes the measurement framing and observed operating frontier, while the production system preserves the infrastructure required to apply that frontier inside governed enterprise decision environments.

\paragraph{Result-set construction} Each reported row is evaluated on the same $226$-example held-out split. Outputs are mapped onto the expected response scale before scoring. The reported modal accuracy and $\Wone$ are point estimates on the scored outputs for that row. Table values are rounded for readability: modal accuracy to three decimals and $\Wone$ to two decimals.

\paragraph{Scoring flow} For each item, the model output is mapped onto the benchmark scale. The modal-accuracy calculation compares the model's most common predicted score with the empirical human modal score. The $\Wone$ calculation compares the model response distribution with the empirical human response distribution through their cumulative distribution functions. Finally, item-level scores are aggregated across the held-out set to produce the table point estimates.

\paragraph{Modal accuracy} For an item with theoretical scale range $[s_{\min}, s_{\max}]$, modal accuracy is:
\begin{equation}
\acct = 1 - \frac{|\hat{s} - s^*|}{s_{\max} - s_{\min}},
\end{equation}
where $\hat{s}$ is the model's most common predicted score and $s^*$ is the most common empirical human score. The term \emph{modal accuracy} is deliberate because the score compares the simulator with the empirical human mode: the behavior selected most often by people. We use it because many enterprise questions start as decision-direction problems: which message, offer, package, or product option moves the population in the intended direction?

\paragraph{Wasserstein-1 distance} For ordinal distributions $\hat{P}$ and $P^*$ with cumulative distribution functions $\hat{F}$ and $F^*$, one-dimensional Wasserstein-1 distance is:
\begin{equation}
\Wone(\hat{P}, P^*) = \int_{-\infty}^{\infty} |\hat{F}(x)-F^*(x)|\,dx.
\end{equation}
Lower is better. $\Wone$ asks how far the simulated population curve must move to match the empirical human curve. We use it because enterprise decisions depend on allocation across ordered states in addition to the headline direction. A launch forecast can identify the leading concept while still putting too much demand in the wrong segment, overstating enthusiasm, or missing the rejection tail. In that case, the headline direction is right but the operating decision can still fail: budget, sales capacity, pricing exposure, and risk review all depend on the population underneath that headline. $\Wone$ is useful because it penalizes both how much probability mass is misplaced and how far it is misplaced across the scale. Confusing adjacent states is less severe than confusing opposite ends of the scale. Wasserstein distance is a standard optimal-transport measure for comparing distributions~\citep{villani2009,ramdas2017wasserstein}.

\begin{tlscope}{Uncertainty and comparability}
Point estimates in the reported table come from the comparable held-out evaluation. Bootstrap confidence intervals are useful for statistical sensitivity analysis~\citep{efron1994bootstrap}, but the reported table uses the same point-estimate convention for every row. The core result is that modal accuracy and distributional fidelity define different operating regimes.
\end{tlscope}

%====================================================================
\clearpage
\section{Results}
\label{sec:results}

\begin{center}
\centering
\includegraphics[width=\linewidth]{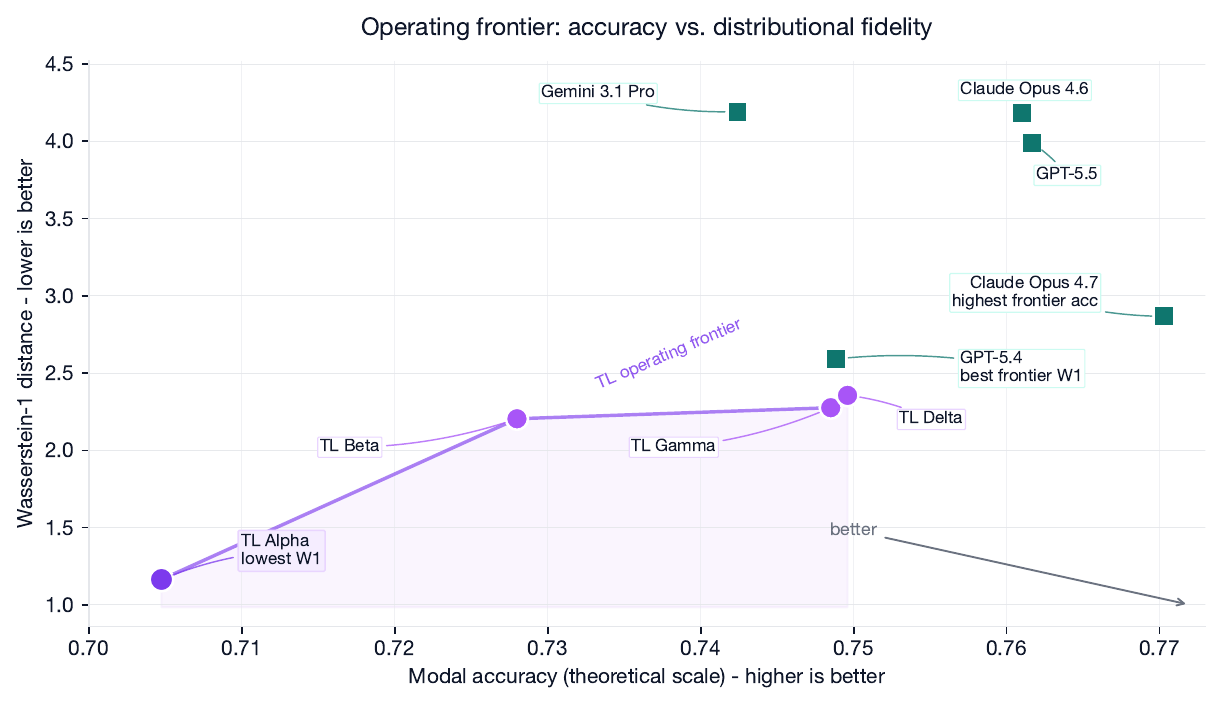}
\captionof{figure}{Operating frontier on the held-out evaluation. Point estimates show the trade-off between modal accuracy and distributional fidelity. Lower $\Wone$ is better; higher modal accuracy is better.}
\label{fig:frontier}
\end{center}

Figure~\ref{fig:frontier} is the paper's central result. It shows behavioral simulation as an operating frontier rather than a single leaderboard. Claude Opus 4.7 has the highest frontier-model modal-accuracy point estimate in the reported result set. GPT-5.4 has the strongest frontier-model $\Wone$ point estimate. TL-Twin Alpha has the lowest observed $\Wone$ overall, while TL-Twin Delta and TL-Twin Gamma sit at point-estimate parity with GPT-5.4 modal accuracy and lower $\Wone$ point estimates. These claims measure different dimensions of the decision problem.

Every row is evaluated on the same 226-example held-out set. The table therefore locates each system on the same modal-accuracy / distributional-fidelity frontier.

\begin{center}
\begin{minipage}{\linewidth}
\captionof{table}{Reported result set. Higher modal accuracy is better; lower $\Wone$ is better. Values are point estimates from the comparable 226-example held-out evaluation.}
\label{tab:results}
\footnotesize
\setlength{\tabcolsep}{4.2pt}
\renewcommand{\arraystretch}{1.04}
\begin{tabular}{@{}p{0.27\linewidth}p{0.40\linewidth}>{\centering\arraybackslash}p{0.15\linewidth}r@{}}
\toprule
System & Role & Modal accuracy & $\Wone$ \\
\midrule
\rowcolor{tlpurplelight}
TL-Twin Alpha & Population-movement model & $0.705$ & $\mathbf{1.16}$ \\
TL-Twin Beta & Calibrated behavior model & $0.728$ & $2.20$ \\
TL-Twin Gamma & Ensemble scenario model & $0.749$ & $2.28$ \\
TL-Twin Delta & Direct scenario model & $0.750$ & $2.36$ \\
\midrule
Claude Opus 4.7 & Frontier model & $\mathbf{0.770}$ & $2.87$ \\
GPT-5.5 & Frontier model & $0.762$ & $3.99$ \\
Claude Opus 4.6 & Frontier model & $0.761$ & $4.18$ \\
GPT-5.4 & Frontier model & $0.749$ & $2.59$ \\
Gemini 3.1 Pro & Frontier model & $0.742$ & $4.19$ \\
\bottomrule
\end{tabular}
\renewcommand{\arraystretch}{1.0}
\end{minipage}
\end{center}

\subsection{Point-estimate deltas}

The distribution-first result is visible directly in the point estimates. TL-Twin Alpha reports $\Wone=1.16$, versus GPT-5.4 at $\Wone=2.59$ and Claude Opus 4.7 at $\Wone=2.87$. On these point estimates, Alpha reduces $\Wone$ by roughly $55\%$ versus GPT-5.4 and roughly $59\%$ versus Claude Opus 4.7. This is the population-movement operating point in the frontier, while TL-Twin Delta and TL-Twin Gamma provide the balanced points.

The balanced TL points tell the second half of the story. TL-Twin Gamma matches GPT-5.4 at the same rounded modal accuracy ($0.749$) while improving $\Wone$ from $2.59$ to $2.28$. TL-Twin Delta reaches $0.750$ modal accuracy with $\Wone=2.36$. These are the operating points for enterprises that need faster scenario execution without giving up distributional measurement.

The frontier-model rows also serve a clear role. Claude Opus 4.7 marks the frontier-model modal-accuracy point estimate in this reported result set. GPT-5.4 marks the strongest frontier-model distributional-fidelity point estimate. GPT-5.5, Claude Opus 4.6, and Gemini 3.1 Pro show how frontier-model accuracy and response-shape fidelity can diverge. That is the core reason the paper reports both axes.

\subsection{Finding 1: distributional fidelity and modal accuracy split}

TL-Twin Alpha is the distributional-fidelity point: $\Wone=1.16$ with modal accuracy $0.705$. Claude Opus 4.7 is the frontier-model modal-accuracy point: modal accuracy $0.770$ with $\Wone=2.87$. GPT-5.4 is the strongest frontier-model distributional-fidelity comparison at $\Wone=2.59$. A single scalar leaderboard would obscure the fact that these systems solve different parts of the decision problem.

This split is the operating frontier. An enterprise decision-maker often knows which error is expensive. If the question is ``which message wins on average?'', modal accuracy matters more. If the question is ``how does the response distribution move across segments?'', $\Wone$ matters more.

\subsection{Finding 2: accuracy parity is one operating point}

TL-Twin Delta is at point-estimate parity with GPT-5.4 on modal accuracy ($0.750$ vs. $0.749$). TL-Twin Gamma gives a second accuracy-preserving point with lower $\Wone$ ($2.28$) at the same three-decimal accuracy level as GPT-5.4 when rounded ($0.749$). The important reading is not a single parity claim. It is that Twinning Labs exposes multiple operating points across accuracy and distributional fidelity while preserving the system boundary enterprises need around memory, governance, and deployment.

This is the strategic value of operating points. Enterprise workflows do not need to become model-selection workflows. The Simulation Engine can route a fast concept screen toward lower-latency scenario execution, route a price-sensitivity or launch-risk workflow toward population-shape fidelity, and record which operating point was used, why it was routed there, and how the resulting distribution moved when the scenario changed.

\subsection{Finding 3: the system boundary changes the comparison}

For a benchmark comparison, a frontier model and a TL operating point are both rows in a table. For enterprise deployment, they are different system objects. A frontier model is a model endpoint in the customer's architecture; Posterior Twins combine governed memory, TL model operating points, digital twin state, scenario orchestration, distributional measurement, and governance. That architectural distinction changes which benchmark evidence can become reusable decision infrastructure.

The important systems question is whether the organization can run a decision portfolio repeatedly, preserve the assumptions behind each run, compare distributions across alternatives, and close the loop when observed outcomes arrive. That is the system boundary around the benchmark result.

\begin{figure}[!htbp]
\centering
\includegraphics[width=\linewidth]{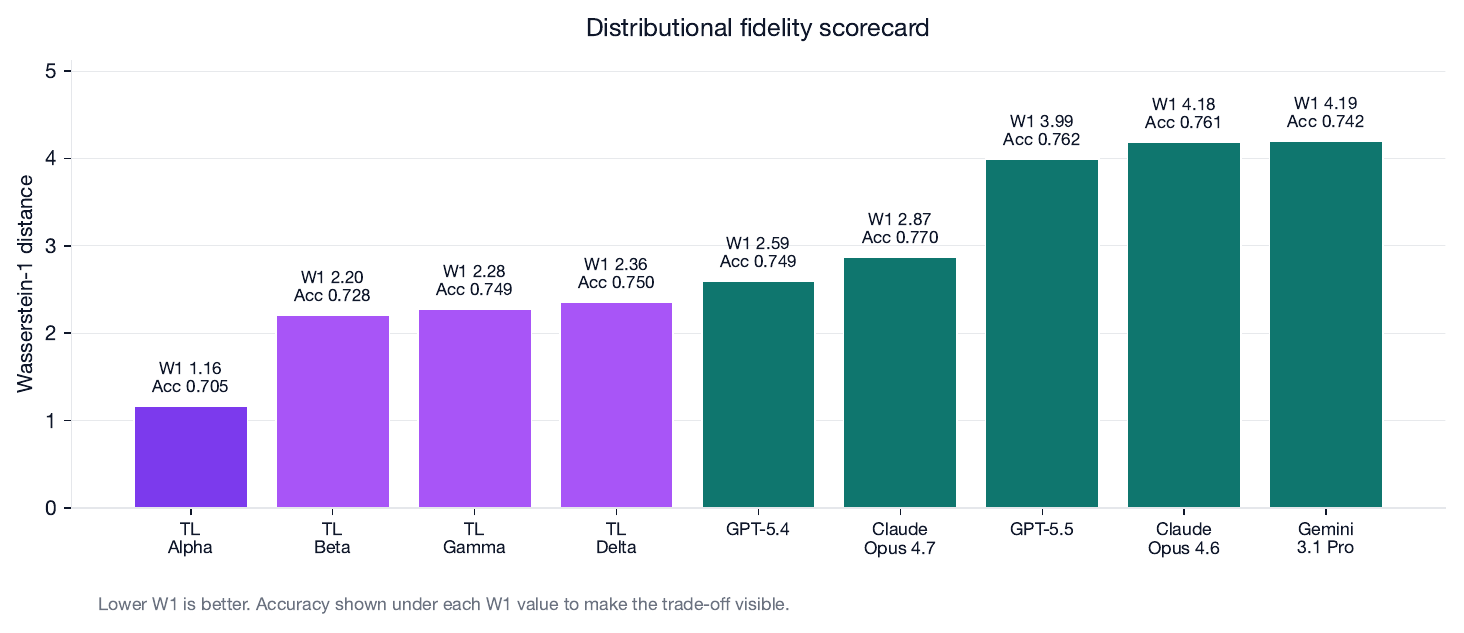}
\caption{Distributional-fidelity scorecard for measured operating points. The scorecard includes modal accuracy under each bar to keep the trade-off visible.}
\label{fig:scorecard}
\end{figure}

\FloatBarrier

%====================================================================
\section{Memory, Digital Twins, Simulation, and Economics}
\label{sec:system}

The benchmark establishes the operating frontier. The system turns that frontier into infrastructure. Posterior Twins need four assets working together: governed memory, memory-grounded digital twins, a simulation engine, and a decision record that can be reused after the run.

\paragraph{Memory Layer} Useful behavioral simulation starts with a connected first-party evidence state. The Memory Layer links the systems where enterprise behavior is observed and updated: customer and account context, product usage, research, campaigns, commercial motion, customer-success interactions, transactions, telemetry, and outcomes. It turns those signals into governed behavioral memory that remains available to twins and scenario runs through abstractions that preserve lineage, permissions, and auditability.

\paragraph{Digital Twin Layer} A Posterior Twin is the actor inside the system. Governed evidence supplies context. The Simulation Engine routes the twin through the relevant TL-Twin operating point. Together, they produce a memory-grounded twin that can choose, respond, act, or generate a behavioral trace under the scenario context.

\paragraph{Simulation Engine} The Simulation Engine turns twins into decision runs. It defines scenario environments, queries the Memory Layer for governed context, routes population-scale simulations across TL-Twin operating points, maps outcomes to the scenario contract, aggregates distributions when the decision depends on population movement, and records the uncertainty context of each result.

\paragraph{Trust boundary} Customer-cloud deployment remains the governance boundary around the system. Sensitive decision evidence often sits in warehouses, CRMs, research repositories, and product analytics systems. The point is not deployment theater; it is keeping governed customer evidence, simulation runs, assumptions, and decision records inside the environment where enterprise teams can actually use them.

\paragraph{Population-scale economics} A behavioral simulator becomes economically useful when teams can run it repeatedly. The unit of work moves from a single model call to a scenario portfolio: owned context, memory-grounded twins, reusable assumptions, traceable evidence, and repeatable measurement. That is why the Memory Layer and Simulation Engine matter commercially. They turn a benchmarked model frontier into a compounding decision layer.

\subsection{Simulation contract}

A clear way to describe the system is as a simulation contract. The contract begins with an \textbf{evidence state}: governed customer, account, product, research, commercial, telemetry, and outcome context assembled from the enterprise systems that maintain it. It then instantiates the relevant \textbf{digital twin population}: memory-grounded actors routed through TL-Twin operating points. It then defines a \textbf{scenario specification}: the decision question, population slice, output target, constraints, assumptions, and alternatives. The run queries the Memory Layer for governed context, executes scenarios across the twin population, maps outcomes to the decision contract, and returns the appropriate decision artifact.

The returned artifact contains the routed operating point, result distribution or generated trace, scenario assumptions, and enough context for review. This gives an enterprise team something it can compare across scenarios, attach to a decision record, and revisit when observed outcomes arrive.

\begin{tlevidence}{System architecture}
Posterior Twins package four assets enterprises already need for serious behavioral simulation: a connected governed evidence state, memory-grounded twins, a repeatable scenario engine, and measured decision artifacts. The benchmark matters because it identifies TL-Twin operating points that the system can route inside scenario execution.
\end{tlevidence}

%====================================================================
\clearpage
\section{Conclusion: Behavioral Decision Infrastructure}
\label{sec:conclusion}

This paper contributes a measurement and systems result: behavioral simulation should be treated as an operating frontier between decision-direction accuracy and distributional fidelity. Once that frontier is visible, a single leaderboard score stops being the right way to evaluate the category.

Twinning Labs contributes an orchestrated behavioral-model family for enterprise simulation. In the comparable reported result set, the TL-Twin Alpha operating point achieves the lowest observed $\Wone$ overall ($1.16$ versus $2.59$ for GPT-5.4, the strongest frontier-model $\Wone$ row). TL-Twin Delta and TL-Twin Gamma preserve near-frontier modal accuracy while also producing lower distributional distance than GPT-5.4. The result is a practical systems advantage: the Simulation Engine routes memory-grounded twins across Alpha, Beta, Gamma, and Delta according to scenario dynamics, evidence state, and decision objective.

That is why the Posterior Twin is the right primitive. A persona prompt produces an answer. A Posterior Twin represents an updated distribution over likely behavior under a decision context. The Memory Layer supplies governed evidence. The TL-Twin model family supplies behavioral operating points. The Simulation Engine supplies scenario environments, automatic routing, population-scale execution, outcome mapping, and distributional aggregation. The resulting artifact is not just text; it is a decision record with a scenario, evidence state, routed operating point, distribution, generated trace where needed, and assumptions that can be reviewed.

\begin{tlcallout}{The category shift}
Posterior Twins sit at the intersection of AI-moderated research, digital-twin systems, multi-agent population simulators, and customer-data decisioning. What changes is the output: not isolated responses or disconnected simulations, but governed decision evidence built from customer memory, calibrated population behavior, and repeatable scenario runs. For enterprises, that means comparing decisions before committing budget, product changes, or market risk.
\end{tlcallout}

For enterprises, this changes the unit of work. The output becomes a repeatable scenario run over memory-grounded twins. Teams compare scenarios, preserve assumptions, review the distribution, and close the loop when observed outcomes arrive; the system handles operating-point routing as part of the run. Each run becomes part of the organization's decision memory: evidence, scenario, twin population, routed operating point, result distribution, generated trace, and observed outcome.

In behavioral simulation, the enduring asset is governed memory, a calibrated behavioral model family, a repeatable simulation engine, and measurement discipline that turn customer evidence into reusable decision infrastructure. These assets convert plausible respondent generation into measured, traceable, and reusable behavioral decision evidence.

Twinning Labs makes population-shape fidelity measurable, routable, and operational inside a digital-twin system. Posterior Twins move enterprise AI from generating what a plausible respondent might say to simulating how a governed population is likely to move under a decision. The result is behavioral decision infrastructure for enterprise simulation.

%====================================================================
\clearpage
\bibliographystyle{abbrvnat}
\bibliography{references}

\end{document}